\pdfoutput=1

\documentclass[11pt]{article}

\usepackage{acl}

\usepackage{times}
\usepackage{latexsym}

\usepackage[T1]{fontenc}

\usepackage[utf8]{inputenc}

\usepackage{microtype}

\usepackage{inconsolata}

\usepackage{graphicx}
\usepackage{amsmath}
\usepackage{amssymb}
\usepackage{pifont}
\newcommand{\xmark}{\ding{55}}%
\usepackage{enumitem}
\usepackage{booktabs}
\usepackage{csquotes}

\title{Eliciting Uncertainty in Chain-of-Thought to Mitigate Bias against Forecasting Harmful User Behaviors}

\author{
    Anthony Sicilia \qquad Malihe Alikhani \\
    Khoury College of Computer Sciences \\
    Northeastern University \\
    \texttt{sicilia.a@northeastern.edu}
}

\usepackage{amssymb}
\usepackage{lipsum}
\usepackage{booktabs}

\begin{document}
\maketitle
\begin{abstract}
Conversation forecasting tasks a model with predicting the outcome of an unfolding conversation. For instance, it can be applied in social media moderation to predict harmful user behaviors before they occur, allowing for preventative interventions. While large language models (LLMs) have recently been proposed as an effective tool for conversation forecasting, it's unclear what biases they may have, especially against forecasting the (potentially harmful) outcomes we request them to predict during moderation. This paper explores to what extent model uncertainty can be used as a tool to mitigate potential biases. Specifically, we ask three primary research questions: 1) how does LLM forecasting accuracy change when we ask models to represent their uncertainty; 2) how does LLM bias change when we ask models to represent their uncertainty; 3) how can we use uncertainty representations to reduce or completely mitigate biases without many training data points. We address these questions for 5 open-source language models tested on 2 datasets designed to evaluate conversation forecasting for social media moderation.
\end{abstract}

\section{Introduction}
\label{sec:intro}
\begin{figure}
    \centering
\includegraphics[width=\columnwidth]{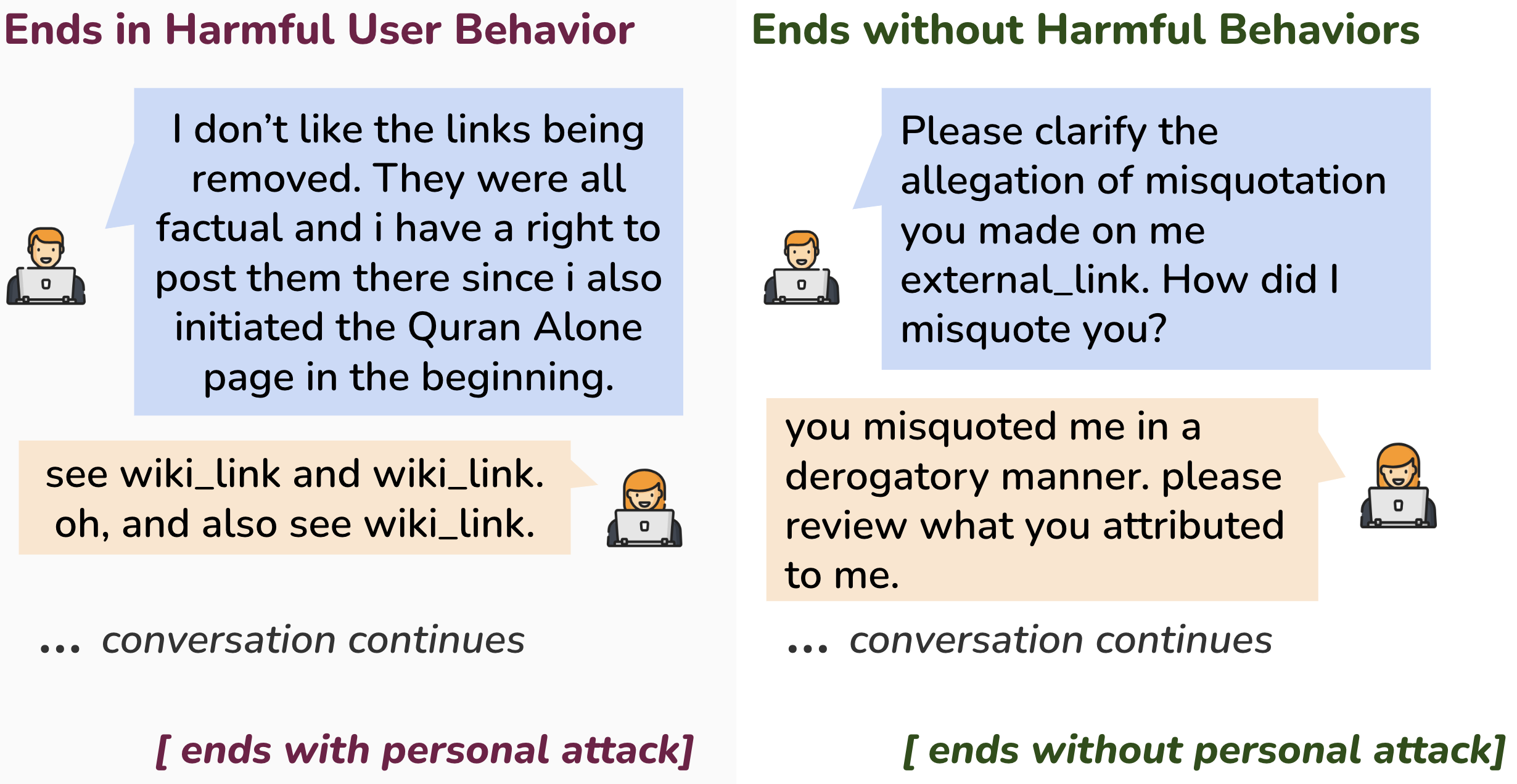}
    \caption{Two difficult social media moderation examples. Both instances appear as if they may derail, leading to harmful user behaviors. Yet, only one does. These are real examples from the moderation corpora we study, identified using this \href{https://awry.infosci.cornell.edu}{online tool.}}
    \label{fig:enter-label}
\end{figure}

Conversation forecasting -- where a model predicts the outcome of a partial conversation -- is useful across many domains, e.g., see research on negotiation dynamics \cite{sokolova-etal-2008-telling}, mental health monitoring \cite{cao2019observing}, and social media moderation \cite{zhang-etal-2018-conversations}. For instance, in online moderation, the forecasting task may be to predict whether a harmful behavior (like digital bullying) will eventually occur in an unfolding conversation, allowing moderators to intervene to prevent these behaviors. Recently, \citet{sicilia2024deal} demonstrate pre-trained language models are relatively effective conversation forecasters, setting themselves apart because they do not require copious amounts of domain-specific training data prior to inference time. Yet, it remains unclear what biases these systems may hold, especially in digital media contexts, where they are specifically asked to predict outcomes that may be harmful to the parties involved (see Figure~\ref{fig:enter-label}).

Indeed, the data used in common instruction-tuning algorithms -- e.g., RLHF \cite{ouyang2022training} and DPO \cite{rafailov2024direct} -- are designed to align language models with human values, and subsequently, avoid any propagation of harm. Meanwhile, the motivating tasks of this paper draw a fine line between propagation and prediction. Surely, ``predicting'' a harmful outcome is not ``speaking into existence'' but it's unclear whether this distinction is lost on ``aligned'' language models. Or, if it is not lost, whether underlying data bias (i.e., against harmful outcomes) pre-disposes language models to propagate this bias when forecasting harmful outcomes. 

While the role of alignment mechanisms in producing model bias is difficult to confirm,\footnote{For instance, pre-training data could also play a role.} our own empirical results, and those of previous work \citep{sicilia2024deal}, indicate current language models are indeed biased against predicting harmful outcomes. Aptly, this paper is interested in mitigating these biases, and we approach this task using uncertainty estimation.

By its nature, conversation forecasting is a highly uncertain task. For instance, two seemingly similar conversations can end with opposite outcomes (e.g., a personal attack vs. an amicable resolution, as in Figure~\ref{fig:enter-label}). While modeling this uncertainty has independent motivations besides the study of bias \citep{sicilia2024deal}, we are specifically interested in how considering uncertainty effects the ``reasoning'' process of language models.\footnote{We do not intend to imply that language models conduct any human-like forms of reasoning. Yet, changing prompts to elicit focus on uncertainty innately changes the tokens on which we condition language model outputs; this is the statistical process which we intend to study.} Indeed, neuroscience (both cognitive and computational) recognizes the role uncertainty plays in human decision-making, wherein the brain is understood to both predict and process different forms of uncertainty \cite{bland2012different}. We hypothesize language models may benefit from utilizing similar patterns of reasoning, having learned these (statistical) patterns from the human-generated text on which they are trained. In particular, we hypothesize elicitation of uncertainty can mitigate bias in model predictions. 

In studying this broader hypothesis, we focus on three central research questions:
\begin{enumerate}[nolistsep]
    \item how does the forecasting accuracy of a language model change when it is prompted to reflect uncertainty in it's prediction;
    \item how does the bias of a language model's forecasts (i.e., against harmful outcomes) change when it is prompted to reflect uncertainty;
    \item and, how can we use a language model's predicted uncertainty to mitigate any such biases.
\end{enumerate}
We address these questions for 5 open-source language models tested on two datasets from the conversation forecasting corpora proposed by \citet{zhang-etal-2018-conversations}, specifically tailored towards harmful behaviors (i.e., personal attacks) in social media.
\section{Background}
\label{sec:back}
\subsection{Conversation Forecasting Setup}
We work within the conversation forecasting framework established by \citet{sicilia2024deal}, wherein the model is tasked with predicting a conversation's outcome. For instance, it may need to predict whether a personal attack will occur (or not). Since the conversation provides only a limited glimpse into the underlying reality, unknown factors like future developments or unobservable mental states introduce an element of randomness, making it challenging to determine the outcome with certainty based solely on the available information.
\paragraph{Task} For a set of natural language tokens $\mathcal{T}$, we assume observation of a partial multi-party dialogue $D \in \mathcal{T}^*$ consisting of $K$ turns. Following \citet{sicilia2024deal}, the length $K$ is a uniform random number between $2$ and the full dialogue length, simulating the ``partial'' property of the dialogue.\footnote{Turns are marked by unique token sequences; e.g., ``Speaker 4: ...''}
These conversations
appear unfinished to the model, but in reality, have an eventual ground-truth outcome $O \in \{0,1\}$, indicating whether a personal attack occurs or does not occur. The task of the model is to predict $O$ given $D$ -- that is, to predict whether a personal attack will occur given the partial conversation.
\paragraph{Metrics} \citet{sicilia2024deal} evaluate the quality of a model's uncertainty estimates when conversation forecasting (i.e., using a metric called the Brier score). We focus on different evaluation metrics, selected to properly answer our distinct research questions. Given a model prediction $\hat{O}$ for $O$, we evaluate the model using the \textbf{accuracy} of the prediction: $\mathbf{E}[\hat{O} = O]$. Besides accuracy, we also report the \textbf{F1 score} to capture both precision and recall. To measure the bias of the predictions, we report the \textbf{statistical bias}:  $\mathbf{E}[\hat{O} - O]$, which is traditional measure of systematic error in an estimator. Specifically, this captures the average trend of the model's errors: whether it \textit{over}-estimates (bias is positive) or \textit{under}-estimates (bias is negative) on average. This type of bias is seemingly different from common quantitative notions of social bias in a model's outputs; e.g., see \citet{gallegos2024bias}. In reality, this (older) measure of bias is a special case of \textit{accuracy parity} \citep{zhao2022inherent} where the group trait of interest, or ``protected attribute,'' is the occurrence of a personal attack. 
\paragraph{Corpora} We consider two corpora in this work:  
\begin{enumerate}[nolistsep]
    \item (\texttt{wiki}) a corpus of conversations from Wikipedia's \textit{talk} page, proposed by \citet{zhang-etal-2018-conversations}, in which authors discuss edits to Wikipedia articles; and 
    \item (\texttt{reddit}) a corpus of conversations from the subreddit ChangeMyView, proposed by \citet{chang-danescu-niculescu-mizil-2019-trouble}, in which redditors try to convince each other to change their position on an (often contentious) issue.
\end{enumerate}
Both corpora come with labels of whether a personal attack eventually occurs. The portion of each dataset we use in this paper contains 100 instances without a personal attack and 100 instances with a personal attack, following the (nearly) even distribution of positive/negative instances in the original data. The average number of tokens in each dataset are 387 and 624, respectively; this is checked \textit{after} we prune turns to simulate partial conversations.
\subsection{Other Related Work}
\paragraph{Conversation Forecasting}
As noted, \citet{zhang-etal-2018-conversations} and \citet{chang-danescu-niculescu-mizil-2019-trouble} provide early investigations and data for forecasting personal attacks during dialogue to proactively moderate online forums. Using the same data, \citet{kementchedjhieva2021dynamic, altarawneh-etal-2023-conversation} propose new models, capitalizing on temporal and social aspects of dialogue.
Meanwhile, forecasting of other conversation outcomes includes task-success \citep{walker2000learning, reitter-moore-2007-predicting}, mental health codes \citep{cao-etal-2019-observing}, emotions \citep{wang-etal-2020-sentiment, matero2020autoregressive}, situated actions \citep{lei-etal-2020-likely}, and financial events \citep{koval-etal-2023-forecasting}. Among these, our work is uniquely positioned by its focus on the relationship between uncertainty and bias when using modern language models for this task. Broadly, studying how language models perform at this task is an important research direction because they promise a pipeline that requires very limited labeled data relative to other, previous directions of study. At the same time, these pre-trained models may have unknown biases, calling for the direction of study proposed in the current paper.
\paragraph{Uncertainty Estimation with LMs}
Modern ``aligned'' language models have been shown to be capable at representing uncertainty in their responses to factual queries, even with minimal supervision \citep{kadavath2022language}. Meanwhile, uncertainty has also been well studied in models without alignment to human preferences \citep{desai-durrett-2020-calibration, jiang2021can, dan-roth-2021-effects-transformer, kong-etal-2020-calibrated, zhang-etal-2021-knowing, li-etal-2022-calibration}
Unlike existing work, ours is interested in how fine-tuning for alignment to human preferences might bias the model against predicting adverse outcomes. As far as how we extract uncertainty estimates from the language model, our work is most in line with that of \citet{lin2022teaching, mielke2022reducing, tian2023just} who all suggest ``direct forecasts'' or uncertainty estimates directly specified in the sampled tokens of the model. These estimates are considered best out-of-the-box for the types of models we study \citep{sicilia2024deal}.

\section{Methods}
\label{sec:meth}
\subsection{Forecasting with Language Models}
Here, we describe prompts used to elicit conversation forecasts. A full example is in the Appendix.
\paragraph{Traditional CoT Classification} To predict conversation outcomes with language models, we simply provide the language model with the partial conversation segment and prompt the language model to predict the outcome. There are some key components to precisely detail our strategy.
\begin{enumerate}[nolistsep, leftmargin=*]
    \item \textbf{Role Play}: As part of the system prompt, we give the language model a ``name'' and ``skill set'' to direct the language model to mimic a task expert. This is a common prompt engineering technique. We use a similar role description as \citep{sicilia2024deal}, emphasizing skills like Theory of Mind and the ability to predict actions/thoughts of different interlocutors.
    \item \textbf{Output Format}: To conclude the system prompt, we direct the model to use an easy-to-parse format; e.g., \texttt{ANSWER = 1} for $O=1$.
    \item \textbf{Context}: To start the user prompt, we explain the context of the conversation; e.g., ``The speakers are discussing edits to a Wikipedia article.'' We then provide context for predicting this specific instance. These include the partial conversation segment (delimited using special token sequences) and the question of interest. Specifically, we ask ``Will a personal attack occur at the end of the conversation?''.
    \item \textbf{Chain of Thought}: We conclude the user prompt with a chain-of-thought trigger phrase. Specifically, we use ``Let's think step by step, but keep your answer concise (less than 100 words).'' This encourages the model to output reasoning for it's answer and has been shown to improve performance \citep{kojima2022large}.
\end{enumerate}
\paragraph{Uncertainty-Aware CoT Classification}
We use largely the same prompting strategy as traditional classification. Instead of asking for an answer directly, we instruct the model to report it's answer on a 10 point Likert scale where 1 indicates ``not likely at all'' and 10 indicates ``almost certainly.'' After parsing the answer (with the same regular expression), we set $O=1$ if the score is greater than 5. We set $O=0$ otherwise. This allows the model to explicitly consider ``uncertainty'' in it's answer as well as the ``reasoning'' process triggered by the chain-of-thought prompting technique.
\paragraph{Post-hoc Intervention for Bias Mitigation}
Besides our initial hypothesis -- that considering ``uncertainty'' in the inference step may improve chain of thought reasoning and subsequent performance -- outputting certainty in the answer allows us to tune the model's answer to our data source. Rather than data- and compute-expensive fine-tuning of model weights, we suggest \textbf{post-hoc forecast scaling}, which is a variant of Platt Scaling, proposed to improve the forecasts of language models by \citet{sicilia2024deal}. If $\hat{P}$ is the parsed and normalized Likert score (i.e., divided by 10), which signals model uncertainty, we use parameters $\tau$ and $\beta$ to scale:
\begin{equation}
\label{eqn:post-process}
\begin{split}
    \hat{Z} & \leftarrow \log \hat{P} / (1 - \hat{P}) \\
    \tilde{Z} & \leftarrow \hat{Z} / \tau - \beta \\
    \hat{P}_\texttt{new} & \leftarrow 1 / (1 + \exp(-\tilde{Z})).
\end{split}
\end{equation}
$\hat{P}_\texttt{new}$ is then used as the new (normalized) Likert score for confidence; i.e., if $10 \times \hat{P}_\texttt{new} > 5$ we set $O = 1$. Parameters are learned by MLE (n=50), treating $\hat{P}_\texttt{new}$ as likelihood for the ground-truth outcome. While this method is known to improve uncertainty estimates, it's not yet been studied in the current paper's context; i.e., exploring its impact on forecasting accuracy or model bias.
\paragraph{Models} We test these prompting and scaling techniques on Llama 3.1 8B and 70B \citep{llama3modelcard}, Mistral 7B v0.3 and Mixtral 8x22B \citep{jiang2023mistral, jiang2024mixtral}, and Qwen2 72B \citep{yang2024qwen2technicalreport}. All models are instruction-tuned variants. We use the default sampling parameter settings for Llama as provided in the official Llama GitHub repository (temp = 0.6, top p = 0.9). For all other models, we use temp = 0.7 and top p = 1. We access models via the \href{https://together.ai}{together AI API}.
\subsection{Semi-Automated Topic Analysis}
\paragraph{Method}
One aspect we explore empirically is the relationship between a model's forecasting bias and the topic of the conversation. This can give us a more fine-grained view of how a model is biased in the context of social media moderation. We use a semi-automated pipeline to predict topics using a large language model. Specifically, we use Meta's Llama 3.1 405B. Our strategy is as follows:
\begin{enumerate}[nolistsep]
    \item Prompt the language model to provide a noun phrase describing the topic of each instance.
    \item Prompt (the same model) to collect the list of sub-topics into higher-level categories.
    \item Iterate step two if the model misses any sub-topics. This process is accelerated with a programmatic check on the model outputs. We re-prompted (in the same conversation context) to tell the model which noun phrases were left out of the current category list.
    \item Manually inspect the final model-generated categories. To improve the categories, we re-organize, combine, and  remove small categories (less than 10 instances).
    \item Ask the model to analyze it's own (author adjusted) categories and provide descriptions.
\end{enumerate}
\paragraph{Topics}
This process only worked well for the \texttt{reddit} corpus (as manually evaluated by the authors based on diversity and correctness). It produced the following categories (and descriptions):
\begin{itemize}[nolistsep,leftmargin=*]
\item \textbf{Social Issues}: ``This category encompasses a wide range of topics related to social justice, equality, and human rights. It includes discussions on discrimination, feminism, LGBTQ+ rights, racism, and other forms of social inequality. Sub-topics also explore issues related to family and relationships, such as marriage, child abuse, and parental leave.''
\item \textbf{Politics and Law}: ``This category delves into the realm of governance, policy-making, and the legal system. It covers topics such as gun control, immigration, free speech, and electoral politics, as well as issues related to national security, terrorism, and international relations. Sub-topics also examine the role of government, the judicial system, and the relationship between citizens and the state.''
\item \textbf{Economics}: ``This category focuses on the production, distribution, and exchange of goods and services. It includes discussions on trade deficits, minimum wage, labor unions, and regulation, as well as emerging topics like cryptocurrency and digital goods. Sub-topics also touch on social welfare and the economic aspects of family relationships, such as alimony and child support.''
\item \textbf{Health}: ``This category explores topics related to physical and mental well-being, including vaccination, mental health, and substance use. It also covers issues related to healthcare policy, medical ethics, and the intersection of health and society, such as prostitution and sexting laws. Sub-topics also examine lifestyle choices, such as veganism and vegetarianism.''
\item \textbf{Culture and ID}: ``This category examines the complex and multifaceted nature of identity, culture, and society. It includes discussions on cultural identity, feminist terminology, indigenous rights, and the Israeli-Palestinian conflict, among others. Sub-topics also explore the intersection of culture and politics, including the role of historical figures, social movements, and cultural protests.
\item \textbf{Tech and Ent}: ``This category delves into the world of technology, entertainment, and media. It covers topics such as ad blocking, game streaming, journalism, and social media, as well as issues related to censorship, art, and sports. Sub-topics also examine the impact of technology on society, including privacy concerns and the ethics of online behavior.''
\item \textbf{Ethics and Morality}: ``This category grapples with fundamental questions about right and wrong, morality, and ethics. It includes discussions on free will, animal rights, organ donation, and evidence-based reasoning, among others. Sub-topics also explore the nuances of human behavior, including discipline, gift giving, and historical judgment.''
\end{itemize}
Descriptions were judged to be accurate by the authors. The full list of sub-topics and super-topics are in the Appendix, along with key prompts.
\section{Experiments}
\label{sec:results}
\begin{table*}[]
\centering\small
\begin{tabular}{lllllllllllll}
\toprule
& \multicolumn{2}{l}{Llama 3.1 8B} & \multicolumn{2}{l}{Llama 3.1 70B} & \multicolumn{2}{l}{Mistral v0.3 7B} & \multicolumn{2}{l}{Mixtral 8x22B} & \multicolumn{2}{l}{Qwen 72B} & \multicolumn{2}{l}{mean \textbf{ACC}} \\\cmidrule(lr){2-3}\cmidrule(lr){4-5}\cmidrule(lr){6-7}\cmidrule(lr){8-9}\cmidrule(lr){10-11}\cmidrule(lr){12-13}
\textbf{uncertainty} & \xmark               & \checkmark           & \xmark              & \checkmark             & \xmark               & \checkmark              & \xmark              & \checkmark             & \xmark            & \checkmark      & \xmark            & \checkmark   \\\midrule
\texttt{wiki}        & 67.5             & 68            & 64              & 62              & 51.5             & 54               & 53              & 58              & 53.5          & 54.5   & 57.9	& \textbf{59.3 }     \\
\texttt{reddit}      & 58               & 57.5          & 66.5            & 61.5            & 52               & 51.5             & 54              & 59.5            & 43.5          & 48.5      & 54.8	& \textbf{55.7}   \\\midrule
mean \textbf{ACC}   & 62.75	& 62.75	& 65.25	& 61.75	& 51.75 & \textbf{52.75} &	53.5 &	\textbf{58.75}* &	48.5 & \textbf{51.5} & 56.35	& \textbf{57.5} \\\bottomrule      
\end{tabular}
\caption{Accuracy of different models at forecasting personal attacks with (\checkmark) and without (\xmark) uncertainty-aware prompting strategy. Accuracy is reported on a 100pt scale. \textbf{Bold} shows improvement from incorporating uncertainty for model/data averages. An asterisk is used to denote statistically significant results (among the averages).}
\label{tab:acc}
\end{table*}
\begin{table*}[]
\centering\small
\begin{tabular}{lllllllllllll}
\toprule
& \multicolumn{2}{l}{Llama 3.1 8B} & \multicolumn{2}{l}{Llama 3.1 70B} & \multicolumn{2}{l}{Mistral v0.3 7B} & \multicolumn{2}{l}{Mixtral 8x22B} & \multicolumn{2}{l}{Qwen 72B} & \multicolumn{2}{l}{mean \textbf{F1}} \\\cmidrule(lr){2-3}\cmidrule(lr){4-5}\cmidrule(lr){6-7}\cmidrule(lr){8-9}\cmidrule(lr){10-11}\cmidrule(lr){12-13}
\textbf{uncertainty} & \xmark               & \checkmark           & \xmark              & \checkmark             & \xmark               & \checkmark              & \xmark              & \checkmark             & \xmark            & \checkmark   & \xmark            & \checkmark        \\\midrule
\texttt{wiki}        & 0.692	& 0.698	& 0.621	& 0.6 &	0.185 &	0.258 &	0.266	& 0.4	& 0.243	& 0.305 & 0.401	& \textbf{0.452}\\
\texttt{reddit}      & 0.702	& 0.699	& 0.747	& 0.712	& 0.461	& 0.497	& 0.494	& 0.61	& 0.199	& 0.383   & 0.521	& \textbf{0.580}   \\\midrule
mean \textbf{F1}   & 0.697	& 0.699	& 0.684	& 0.656	& 0.323	& \textbf{0.378}	& 0.38	& \textbf{0.505}	& 0.221	& \textbf{0.344} & 0.461	& \textbf{0.516}\\\bottomrule      
\end{tabular}
\caption{F1 scores of different models at forecasting personal attacks with (\checkmark) and without (\xmark) uncertainty-aware prompting strategy. F1 ranges from 0 to 1. \textbf{Bold} shows improvement from incorporating uncertainty.}
\label{tab:f1}
\end{table*}
In general, we use Hoeffding's Inequality to test statistical significance at level $\alpha = 0.05$. It provides a versatile (albeit, conservative) confidence interval with limited assumptions, making it applicable to accuracy (\textbf{ACC}) \textit{and} statistical bias (\textbf{SB}).
\subsection{Uncertainty and Forecasting Performance}
\begin{figure}
    \centering
    \includegraphics[width=\columnwidth]{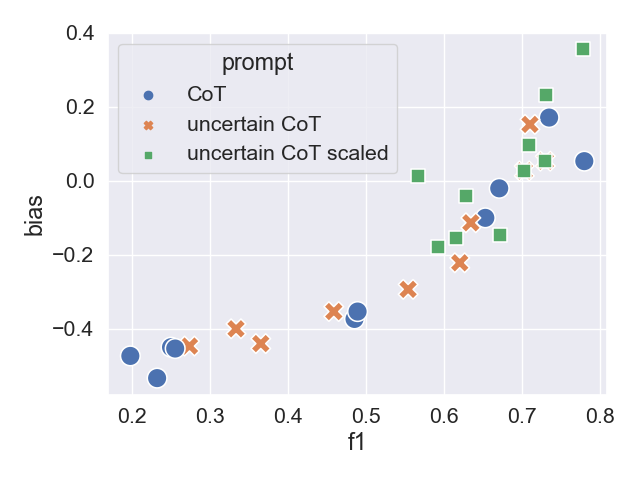}
    \caption{F1 v. Bias for all models / datasets with different inferences strategies. CoT refers to our standard conversation forecasting prompt (i.e., which uses CoT), while uncertain CoT ask the model to represent it's uncertainty in place of direct classification. Scaling refers to post-hoc scaling and is only applicable to the former strategy. It is best to have near 0 bias and high F1 score.}
    \label{fig:scatter}
\end{figure}
\begin{displayquote}
\textit{RQ1: How does uncertainty-aware inference impact the forecasting performance of language models?} \\
\textit{A: Some language models, especially those that perform poorly initially, benefit from considering uncertainty.}
\end{displayquote}
\paragraph{Forecasting Accuracy Results} Table~\ref{tab:acc} shows forecast accuracy across models and datasets with and without the uncertainty-aware prompt strategy. For 3 out of 5 models, the uncertainty-aware strategy leads to improved performance on average. Average increases in accuracy range from 1\% up to 5.25\%, which on our dataset corresponds to about 3 to 13 more correct predictions, respectively. The Llama 3.1 series (8B and 70B) are the only models which do no benefit from the uncertainty-aware strategy. For the 8B model, performance is unchanged (averaged across datasets). For the 70B model, performance is reduced by nearly 4\%. For both datasets, the uncertainty-aware strategy lead to improved performance (on average). Average increases are near 1\% for the Wikipedia corpus and the Reddit corpus. The only statistically significant improvement in performance comes when we apply the uncertainty-aware strategy to Mixtral.
\paragraph{Forecasting F1 Results} Table~\ref{tab:f1} shows F1 scores for forecasts across models and corpora. When considering precision and recall of inferences (F1 is their harmonic mean), we find results are largely consistent with those reported for accuracy. Three of five models show improvement, meanwhile both datasets show improvement. Relative performance of models is also consistent: Qwen2 does worst, is improved by the Mistral models, and further improved by the Llama 3.1 series.
\paragraph{Discussion} Findings indicate that considering uncertainty in the LM forecast either has little impact (on average) or a slight positive one, for certain models. One observation is that the best performing models (the Llama 3 series) are either unaffected by the change in prompt (in case of the 8B model) or negatively effected by the prompt (in case of the 70B model). Although, the negative result is not statistically significant. We hypothesize a saturation effect may occur for these high performing models, where there is little additional predictive power to be gained through simple means like prompt engineering. Comparing these results to related literature suggests this may be the case. Indeed, in a similar experimental setup (albeit, slightly easier) an average accuracy near 64\% is achieved by a specialized model \textit{which is trained on the dataset} \citep{altarawneh-etal-2023-conversation}, showing (potentially) that waning amounts of insight can be gained on this highly uncertain task once accuracy reaches a certain threshold. On the other hand, for models with a worse baseline accuracy, considering uncertainty in the prompt does seem to offer some benefit to the inference process. As we note previously, we hypothesize this is due the interaction between the chain-of-thought ``reasoning'' and the answer-format (which represents model uncertainty). Considering uncertainty may tap into patterns of ``reasoning'' learned from the training data that are overall beneficial.
\begin{table*}[]
\centering\small
\begin{tabular}{lllllllllllll}
\toprule
& \multicolumn{2}{l}{Llama 3.1 8B} & \multicolumn{2}{l}{Llama 3.1 70B} & \multicolumn{2}{l}{Mistral v0.3 7B} & \multicolumn{2}{l}{Mixtral 8x22B} & \multicolumn{2}{l}{Qwen 72B} & \multicolumn{2}{l}{mean \textbf{SB}} \\\cmidrule(lr){2-3}\cmidrule(lr){4-5}\cmidrule(lr){6-7}\cmidrule(lr){8-9}\cmidrule(lr){10-11}\cmidrule(lr){12-13}
\textbf{uncertainty} & \xmark               & \checkmark           & \xmark              & \checkmark             & \xmark               & \checkmark              & \xmark              & \checkmark             & \xmark            & \checkmark & \xmark            & \checkmark          \\\midrule
\texttt{wiki}      & -0.03	& 0.01	&  -0.12	&  -0.12	&  -0.48& 	-0.45	& -0.44	& -0.37	& -0.46	& -0.42 & -0.30	& \textbf{-0.27} \\
\texttt{reddit}      & 0.21	&0.20	&0.10	&0.11	&-0.34	&-0.27	&-0.32	&-0.20	&-0.53	&-0.40   & -0.18	& \textbf{-0.11}   \\\midrule
mean \textbf{SB}   & 0.09	&0.11	& -0.01	& -0.01	& -0.41	&\textbf{-0.36}	& -0.38	&\textbf{-0.29}*	&-0.49	&\textbf{-0.41}* & -0.24	& \textbf{-0.19} \\\bottomrule      
\end{tabular}
\caption{Statistical bias of models forecasting personal attacks with (\checkmark) and without (\xmark) uncertainty-aware prompting strategy. \textbf{SB} ranges between -1 and 1 with closer to 0 being best. \textbf{Bold} shows improvement from incorporating uncertainty. An asterisk is used to denote statistically significant results (among the averages).}
\label{tab:bias}
\end{table*}
\begin{figure*}
    \centering
    \includegraphics[width=\textwidth]{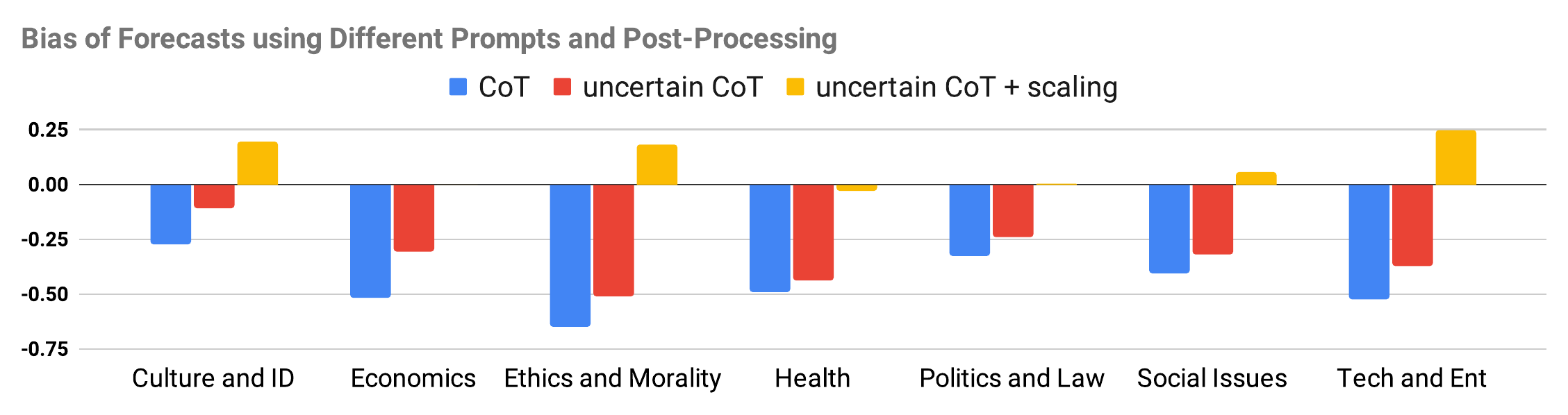}
    \caption{Statistical Bias of Forecasts on Reddit for Mistral models and Qwen2. Language models either use uncertainty estimates to report inferences (uncertain CoT) or make traditional binary decsions (CoT). Impact of post-hoc scaling is also shown for the former of these methods. Topics are determined using the method from \S~\ref{sec:meth}.}
    \label{fig:topics}
\end{figure*}
\subsection{Uncertainty and Forecasting Bias}
\begin{displayquote}
\textit{RQ2: How does uncertainty-aware inference impact forecasting bias?} \\
\textit{A: While some language models consistently under-predict the occurrence of personal attacks, considering uncertainty is able to partially reduce this bias.}
\end{displayquote}
\paragraph{Forecasting Bias Results} Table~\ref{tab:bias} shows statistical bias of language model forecasts with and without consideration of uncertainty at inference-time. Uncertain inferences reduce bias for three out of five models. Again, the Llama 3.1 series are the only models that do not show any benefit. In this case, bias is relatively consistent with/without uncertainty (unlike the drop in accuracy for the 70B model observed in Table~\ref{tab:acc}). Bias was often negative, \textit{indicating that models typically under-predict the occurrence of a personal attack}; i.e., on average, they predict no personal attack when an attack does in fact occur. Only the Llama 3.1 series showed any sign of positive bias (specifically, on the Reddit corpus). Reductions in bias range from 0.05 up to 0.09. In our context, this means use of uncertainty corrected 5 out of 100 or 9 out of 100 false negatives, respectively. For some models (Mixtral and Qwen2), this reduction is statistically significant. Both datasets also experience reduction in bias on average, with 3 out of 100 and 7 out of 100 less false negatives for the Wikipedia corpus and Reddit corpus, respectively. These reductions were not statistically significant.
\paragraph{Interactions Between Topic and Bias} Figure~\ref{fig:topics} shows the relationship between bias and different topics identified using the method from \S~\ref{sec:meth} applied to the Reddit corpus. We limit consideration to the Mixtral models and Qwen2, since these models exhibited consistent negative bias (i.e., systematic under-prediction of personal attacks). For traditional inference without uncertainty (traditional CoT), bias is most prominent on Reddit conversations about ``Ethics and Morality'' followed by conversation about ``Economics'' or ``Tech and Entertainment.'' When uncertainty is considered during inference (uncertain CoT), bias is reduced for all topics. One of the biggest reductions occurs for the ``Economics'' topic. For both forecasting methods, the topics with the lowest overall bias are ``Culture and Identity'' and ``Politics and Law.''
\paragraph{Discussion} Findings indicate that most language models exhibit negative statistical bias (systematic under-prediction) when forecasting personal attacks. This lends evidence to our over-arching hypothesis -- that AI alignment mechanisms can bias language models against predicting harmful outcomes -- since under-prediction of a personal attack is indeed a harmful outcome. Of course, it is difficult to confirm this idea without transparent access to training data and methods (for alignment) as well instruction-tuned models, which are guaranteed to be ``un-aligned'' along the dimensions of interest. In any case, findings also indicate that uncertainty-aware inference with language models is able to reduce negative bias. As before, the impact of uncertainty-aware inference is not consistent across models: the more biased models experience the greatest degrees of bias reduction. For two models, this reduction was even statistically significant. We hypothesize the disparity across models again may be due to a saturation effect, as models which are not consistently biased do not have consistent patterns of ``reasoning'' that can be modified by consideration of uncertainty. We also observe that bias is not uniform across topics, nor is bias reduction (by uncertain CoT). We do not find any consistent properties among topics, which cause more/less bias. Yet, if our overarching hypothesis is correct -- that AI alignment is a cause of bias -- then this non-uniformity may be related to the types/amounts of data used during alignment.
\begin{table*}[]
\centering\small
\begin{tabular}{lllllllllllll}
\toprule
& \multicolumn{2}{l}{Llama 3.1 8B} & \multicolumn{2}{l}{Llama 3.1 70B} & \multicolumn{2}{l}{Mistral v0.3 7B} & \multicolumn{2}{l}{Mixtral 8x22B} & \multicolumn{2}{l}{Qwen 72B} & \multicolumn{2}{l}{mean \textbf{F1}} \\\cmidrule(lr){2-3}\cmidrule(lr){4-5}\cmidrule(lr){6-7}\cmidrule(lr){8-9}\cmidrule(lr){10-11}\cmidrule(lr){12-13}
\textbf{scaling} & \xmark               & \checkmark           & \xmark              & \checkmark             & \xmark               & \checkmark              & \xmark              & \checkmark             & \xmark            & \checkmark   & \xmark            & \checkmark        \\\midrule
\texttt{wiki}        & 0.703	& 0.703	& 0.634	& 0.628	& 0.274	& 0.593	& 0.459	& 0.671	& 0.333	& 0.731	& 0.481	& \textbf{0.665} \\
\texttt{reddit}      & 0.710	& 0.709	& 0.730	& 0.730	& 0.554	& 0.567	& 0.620	& 0.779	& 0.365	& 0.615	& 0.596	&  \textbf{0.680}   \\\midrule
mean \textbf{F1}   & 0.707	& 0.706	& 0.682	& 0.679	& 0.414	& \textbf{0.580}	& 0.539	& \textbf{0.725}	& 0.349	& \textbf{0.673}	& 0.538	& \textbf{0.673}\\\bottomrule      
\end{tabular}
\caption{F1 scores of different models with (\checkmark) and without (\xmark) post-hoc scaling; i.e., so all models are prompted to express uncertainty. Post-hoc scaling uses a 50 sample dev. set and results are reported on remaining (held out) data. \textbf{Bold} shows improvement from incorporating uncertainty for model/data averages.}
\label{tab:f1_scale}
\end{table*}
\begin{table*}[]
\centering\small
\begin{tabular}{lllllllllllll}
\toprule
& \multicolumn{2}{l}{Llama 3.1 8B} & \multicolumn{2}{l}{Llama 3.1 70B} & \multicolumn{2}{l}{Mistral v0.3 7B} & \multicolumn{2}{l}{Mixtral 8x22B} & \multicolumn{2}{l}{Qwen 72B} & \multicolumn{2}{l}{mean \textbf{SB}} \\\cmidrule(lr){2-3}\cmidrule(lr){4-5}\cmidrule(lr){6-7}\cmidrule(lr){8-9}\cmidrule(lr){10-11}\cmidrule(lr){12-13}
\textbf{scaling} & \xmark               & \checkmark           & \xmark              & \checkmark             & \xmark               & \checkmark              & \xmark              & \checkmark             & \xmark            & \checkmark   & \xmark            & \checkmark        \\\midrule
\texttt{wiki}        & 0.03	& 0.03	& -0.11	& -0.04	& -0.45	& -0.18	& -0.35	& -0.15	& -0.40	& 0.23	& -0.26	& \textbf{-0.02}* \\
\texttt{reddit}      & 0.15	& 0.10	& 0.05	& 0.05	& -0.29	& 0.01	& -0.22	& 0.36	& -0.44	& -0.15	& -0.15	& \textbf{0.07}   \\\midrule
mean \textbf{SB}   & 0.09	& \textbf{0.06}	& -0.03	& \textbf{0.01}	& -0.37	& \textbf{-0.08}*	& -0.29	& \textbf{0.11}*	& -0.42	& \textbf{0.04}*	& -0.20	& \textbf{0.03}\\\bottomrule
\end{tabular}
\caption{Statistical bias of different models with (\checkmark) and without (\xmark) post-hoc scaling; i.e., so all models are prompted to express uncertainty. Post-hoc scaling uses a 50 sample dev. set and results are reported on remaining (held out) data. \textbf{Bold} shows improvement from incorporating uncertainty for model/data averages.}
\label{tab:bias_scale}
\end{table*}
\subsection{More Benefits of Uncertainty: Scaling}
\begin{displayquote}
    \textit{RQ3: Can post-hoc scaling of uncertainty estimates further mitigate bias without impacting accuracy?} \\
    \textit{A: Yes. Scaling consistently produces the least biased and most accurate forecasts.}
\end{displayquote}
\paragraph{Forecasting Accuracy Results} Table~\ref{tab:f1_scale} shows F1 scores for language model forecasts with and without post-hoc scaling of uncertainty estimates. Note, this implies we use the uncertain CoT strategy, since scaling is not possible with traditional CoT. Scaling improves F1 scores by almost 20 pts (out of 100) for Mistral models and more than 30 pts for Qwen2. The Llama 3.1 series remain as the ``odd-models-out'' with their high performance being maintained after the application of scaling. All datasets also show substantial improvements in F1 score after application of scaling.
\paragraph{Forecasting Bias Results} Table~\ref{tab:bias_scale} shows statistical bias with and without post-hoc scaling. Scaling is able to reduce the magnitude of bias for all models, including three (out of five) statistically significant reductions (i.e., all models except the Llama 3.1 series). Average reduction in bias across datasets is also consistent with statistically significant reduction on the Wikipedia corpus. From Figure~\ref{fig:topics}, we more easily see that scaling tends to lead to slight positive bias (less in magnitude then the original negative bias). 
\paragraph{Interaction Between Forecasting Bias and Accuracy}
Figure~\ref{fig:scatter} shows bias and F1 score simultaneously via a scatter plot, for all models/data, organized by prompt strategy and use of scaling. Reductions in bias generally correlate with improved accuracy (an apparent quadratic relationship). Use of all proposed methods (uncertainty-aware CoT with scaling) creates a unique cluster of data points with near 0 bias and high F1 score.
\paragraph{Discussion} Findings show that using a small amount of data for post-hoc scaling consistently improves both F1 score and bias by a relatively large magnitude. We remark, this is a benefit of using uncertainty estimates to make predictions, since post-hoc scaling is not possible for traditional CoT classification. One interesting point is that the Llama 3.1 series remains relatively unaffected by any of our modifications. Again, we believe this to be an effect of saturated (high) performance out-of-the-box. We can understand why scaling works from a mathematical perspective. In particular, the parameter $\beta$ acts to remove systematic biases from the latent score $\hat{Z}$ in Eq.~\eqref{eqn:post-process}. If latent scores are typically higher than they should be (i.e., leading to higher forecast confidence, and thus, over-prediction), the MLE optimization uses $\beta$ to lower these latent scores systematically across all predictions. We hypothesize the reason this correction sometimes leads to positive bias is from over-fitting to the small data sample used for MLE.
\section{Conclusions}
This paper studies three research questions about the interaction between uncertainty estimation and forecast bias for social media moderation using language models. Briefly, our findings show how asking language models to represent their uncertainty when forecasting personal attacks can reduce bias and increase accuracy, especially if a small amount of data is available to fine-tune these inferences.

One interesting point, which we are unable to address, is the root cause of the biases observed. We speculate this is a result of alignment mechanisms biasing language models against predicting the harmful outcomes we wish to forecast (i.e., personal attacks). Yet, more transparency in language model training is needed to investigate this issue.

\section*{Limitations}
As noted in our conclusions, some key hypotheses of our work remain under-explored. Specifically, the cause of observed biases in the language models we study. Working with open-source language models that have closed-source training pipelines makes this a difficult research question to definitely handle. On the other hand, the research questions we \textit{do} answer may also have limited interpretation outside of the contexts in which we study them; i.e., the specific models and datasets explored in \S~\ref{sec:results}. A compounding issue of our analysis is the relatively small test sets we explore (200 instances, due to paper budget) which limited the statistical power of our study, as highlighted by the relatively few statistical significant results.

\section*{Ethics Statement}
While the focus of this work is on analyzing (and mitigating) the bias of the language models we study, we emphasize that models which employ our proposed techniques still incur some bias. This can have direct, negative impact on users if these models are used for social media moderation in a automated pipeline without appropriate human checks. Even with human checks, if these models are used for decision-making, they may influence their human users in unknown ways, which can have unknown (and vast) negative impacts on online communities where they are deployed. Not to mention, we have only explored a very small subset of the potential biases these pre-trained models can possibly have. Other (social) biases may also exist in these models, which our methods are not explicitly designed to counteract and which can also have negative impacts on (vast) numbers of users if used for semi-automated decision-making. These caveats should be carefully considered and studied before systems like the language models we study are used for any automated moderation decisions.

One additional issue is the broader of role content moderation on the internet, and how decisions in content moderation can broadly impact online discourse. The question of who makes moderation decisions, how these decisions are made, and whether moderation should occur at all are each important issues of social debate, which we do not address in this paper. Tacitly, the datasets we study make some claim about what behaviors should be allowed (or not allowed) on online forums, as annotated by human moderators and crowd-workers. We emphasize these distinctions are for the purpose of research study alone, and the content of this data (used for learning and evaluation) should be carefully considered prior to it's use to make decisions or deploy models in real online communities.

\section*{Acknowledgements} This research was supported in part by Other Transaction award HR0011249XXX from the U.S. Defense Advanced Research Projects Agency (DARPA) Friction for Accountability in Conversational Transactions (FACT) program. Thanks to Jack Hessel for helpful discussion about early versions of this work.

\clearpage

\clearpage
\bibliography{custom}
\clearpage
\appendix
\section{Appendix}
\label{sec:appendix}
\subsection{Forecasting System Prompt Example}
You are TheoryOfMindGPT, an expert language model at using your theory-of-mind capabilities to predict the beliefs and actions of others in human conversations. You will be given an unfinished conversation between two speakers. Put yourself in the mindset of the speakers and try to reason about the requested conversation outcome. Use the keyword "ANSWER" to report your prediction for the outcome of interest. Report your answer on a scale from 1 to 10 with 1 indicating "not likely at all" and 10 indicating "almost certainly". For example, "ANSWER = 7" would mean you think the outcome is fairly likely.
\subsection{Forecasting User Prompt Example}
In the following conversation segment, the speakers are negotiating how to allocate available resources among themselves.
\\ \\
\noindent[SEGMENT START]

\noindent Speaker 0: Hello how are you?\\
\noindent Speaker 1: Hello! I am doing well. How about you? \\
Speaker 0: I'm doing well. I'm trying to prepare for this camping trip. \\
\noindent Speaker 1: Me too. \\
\noindent Speaker 0: What are you looking for?...\\
\noindent [SEGMENT END]
\\ \\
\noindent Now, fast-forward to the end of the conversation. Will both speakers be satisfied at the end of the conversation? Let's think step by step, but keep your answer concise (less than 100 words).

\subsection{Topic Model System Prompt}
You are TopicClassifierGPT, an expert language model at assigning topics to conversations across the internet. Try to categorize the topic of the conversation using only one or two words, so that your categories can be automatically grouped and analyzed later. Topics should be nouns or noun phrases that provide an answer to the question: "What are the speakers discussing?" Use the keyword "ANSWER" to report your predicted category. For example, "ANSWER = Religion" could be used for a conversation that is broadly about religion.

\subsection{Topic Model User Prompt}
In the following conversation segment, \\\\
\noindent ... \{\textit{same as forecasting prompt}\}
\\\\
\noindent [SEGMENT END]
\\ \\
\noindent What is the topic of the conversation?

\subsection{Topics}
\begin{itemize}
    \item "Social Issues": [
        "homophobia", "transgenderism", "transgender issues", "transgender rights", "lgbt rights", "islamophobia", "racism", "sexism", "discrimination", "feminism", "social justice", "equal pay", "body image", "objectification", "rape", "sexual assault", "hate speech", "slurs", "marriage pressure", "alimony", "child support", "parental leave", "child abuse", "bullying", "polygamy"
    ],
    \item "Politics and Law": [
        "politics", "gun control", "immigration ban", "judicial bias", "free speech", "affirmative action", "abortion", "censorship", "media bias", "socialism", "communism vs capitalism", "electoral college", "government", "nationalism", "patriotism",
        "travel ban", "us-saudi relations", "terrorism", "military draft", "war", "nuclear power", "capital punishment", "self-defense", "gun ownership", "gun rights", "gun regulation", "gun violence", "dueling laws", "prison", "corporal punishment", "death penalty", "military spending", "immigration", "don't ask don't tell (dadt)", "immigration enforcement", "immigration policy"
    ],
    \item "Economics": [
        "economics", "cryptocurrency", "digital goods", "trade deficits", "minimum wage", "labor unions", "regulation", "social welfare",
        "alimony", "child support"
    ],
    \item "Health": [
        "mental health", "vaccination", "vaccines", "cannabis", "marijuana", "opium trade", "prostitution", "sexting laws", "necrophilia",
        "veganism", "vegetarianism", "gmos"
    ],
    \item "Culture and ID": [
        "cultural identity", "feminist terminology", "islam", "indigenous rights", "israeli-palestinian conflict", "israel", "jordan peterson", "hillary clinton emails", "donald trump", "trayvon martin case", "kavanaugh nomination", "russian investigation",
        "cults vs religion", "historical figures", "\#metoo movement", "flag protest", "pride", "racial protests", "diversity debate", "transgender identity", "pronouns", "transgender dating", "transgender athletes", "transgender youth", "pride parades", "race genetics"
    ],
    \item "Tech and Ent" : ["ad blocking", "game streaming", "journalism", "media bias", "censorship", "art censorship", 
        "social media", "adblocking", "privacy", "american football", "college football", "sports", "star trek", "transgender athletes"
    ],
    \item "Ethics and Morality": [
        "morality", "ethics", "free will", "circumcision", "animal rights", "organ donation", "evidence", "argumentation", "discipline", "historical judgment", "merging", "gift giving", "tipping", "hunting", "protected classes"
    ]
\end{itemize}

\end{document}